\PassOptionsToPackage{warn}{textcomp}
\documentclass[format=sigconf, review=false, anonymous=false]{acmart}
\usepackage{amsmath}
\usepackage{bbold}
\usepackage{booktabs} 
\usepackage[english]{babel}
\usepackage[utf8]{inputenc}
\usepackage[T1]{fontenc}
\usepackage{times}
\usepackage{tabularx}
\usepackage{subfigure}
\usepackage{relsize}
\usepackage{spverbatim}
\usepackage[smaller,nolist]{acronym}
\usepackage{url}
\usepackage{soul}
\usepackage[subtle]{savetrees}
\usepackage[ruled]{algorithm2e} 
\usepackage{listings}

\newcommand{\wikidisamb}[0]{\textrm{Wiki-Disamb30}}
\newcommand{\wikidatadisamb}[0]{\textrm{Wikidata-Disamb}}
\newcommand{\glove}[0]{\emph{Glove}}
\newcommand{\tensorflow}[0]{\emph{Tensorflow}}
\newcommand{\wikidata}[0]{\emph{Wikidata}}
\newcommand{\wikipedia}[0]{\emph{Wikipedia}}

\begin{acronym}[Bi-LSTM]
    \acro{Bi-GCN}{Bi\--directional Graph Convolutional Network}
    \acro{Bi-LSTM}{Bi\--directional Long Short-Term Memory}
    \acro{LSTM}{Long Short-Term Memory}
    \acro{GCN}{Graph Convolutional Network}
    \acro{GRU}{Gated Recurrent Unit}
    \acro{LSTM}{Long Short-Term Memory}
    \acro{NER}{Named Entity Recognition}
    \acro{NLP}{Natural Language Processing}
    \acro{PoS}{Part\--of\--Speech}
    \acro{RNN}{Recurrent Neural Network}
    \acro{EL}{Entity Linking}
    \acro{NED}{Named Entity Disambiguation}
    \acro{WSD}{Word Sense Disambiguation}
\end{acronym}

\setcopyright{acmcopyright}

\acmDOI{0000001.0000001}

\begin{document}
\title{Named Entity Disambiguation using Deep Learning on Graphs}
 
\author{Alberto Cetoli, Mohammad Akbari, Stefano Bragaglia, Andrew D. O'Harney, Marc Sloan}
\orcid{}
\affiliation{%
  \institution{Contextscout}
  \streetaddress{Monmouth House, 58-64 City Rd}
  \city{London}
  \postcode{EC1Y 2AL}
  \country{UK}}
\email{{alberto, stefano, mohammad, andy, marc}@contextscout.com}
\renewcommand\shortauthors{Cetoli, A. et al}

\begin{abstract}
We tackle \ac{NED} by comparing entities in short sentences with \wikidata{} graphs. 
Creating a context vector from graphs through deep learning is a challenging problem that has never been applied to \ac{NED}. 
Our main contribution is to present an experimental study of recent neural techniques, as well as a discussion about which graph features are most important for the disambiguation task. 
In addition, a new dataset (\wikidatadisamb{}) is created to allow a clean and scalable evaluation of \ac{NED} with \wikidata{} entries, and to be used as a reference in future research.
In the end our results show that a \ac{Bi-LSTM} encoding of the graph triplets performs best, improving upon the baseline models and scoring an \rm{F1} value of $91.6\%$ on the \wikidatadisamb{} test set
\footnote{
The dataset and the code for this paper can be found at \url{https://github.com/contextscout/ned-graphs}
}
.
\end{abstract}

%
%


%
%


\maketitle

\section{Introduction and Motivations} 
\label{sec:introduction_and_motivations}

A mentioned entity in a text may refer to multiple entities in a knowledge base. 
The process of correctly linking a mention to the relevant entity is called \ac{EL} or \ac{NED} \cite{bunescu_using_nodate}. 
Entity disambiguation is different from \ac{NER}, where the system must detect the relevant mention boundaries given a definite set of entity types.
In \ac{NED}, the system must be able to generate a context for an entity in a text and an entity in a knowledge base, then correctly link the two.
\ac{NED} is a crucial step in web search tasks \cite{blanco_fast_2015, artiles_role_2009, cucerzan_large-scale_nodate}, data mining \cite{dorssers_ranking_nodate, chang_comparison_nodate,hoffart_robust_nodate}, and semantic search \cite{meij2014,Dietz:2017:UKG:3018661.3022756}.
Arguably, disambiguation falls into the category of tasks where humans still vastly outperform algorithmic solutions. 
This is what makes research into \ac{NED} so relevant in today's data mining landscape.

A background knowledge base can appear in many forms: 
as a collection of texts, as a relational database, or as a collection of graphs in a graph database. 
Representing data as ensembles of linked information is an increasing popular form of storage. 
One example can be found in the successful \wikidata{} database \cite{Vrandecic2012}, which aims to mirror the content of \wikipedia{} in a linked format.

Both \wikipedia{} and \wikidata{} contain potentially ambiguous entities. 
For example, when searching for information about \emph{Captain Marvel}, the results should depend on the context in which this entity appears.
Indeed, the name \emph{Captain Marvel} is a character from Marvel comics and a nickname for \emph{Michael Jordan}, the basketball player. 
\begin{figure}[ht!]
    \subfigure[][]{%
    \label{fig:cmarvel}
    \centering
        \includegraphics[width=0.22\textwidth]{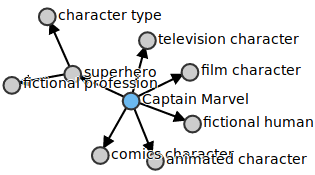}}%
    \qquad
    \subfigure[][]{%
    \label{fig:mjordan}
    \centering
        \includegraphics[width=0.18\textwidth]{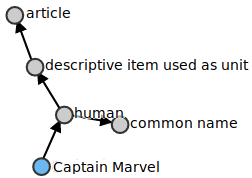}}%
    \caption{The same name can be associated to two different \wikidata{} items. Here we show only the \emph{instance\_of} relationships.}
    \label{fig:graphs}
\end{figure}
With \wikidata{} the ambiguity can be resolved by looking at the information linked to both entities, as we show in Fig. 1. 
More specifically, the issue of disambiguating entities using information in a graph format has been addressed sparsely in the literature. We wish to contribute on this topic with the current paper.

The main contributions of this work are two-fold: 
First, we aim to empirically evaluate different deep learning techniques to create a context vector from graphs, aimed at high-accuracy \ac{NED}. 
This is the most novel aspect of our work, as there is currently no study on a neural approach for entity disambiguation using graphs as background knowledge.
Current state-of-the-art algorithms \cite{raiman_deeptype:_2018} are able to build a context from \wikipedia{} pages, but many academic and commercial projects use a graph-like knowledge base. 
An excellent study on \ac{NED} with graphs has been done in 2014 within \cite{mika_agdistis_2014}, when the techniques for using neural networks on graphs were still under-developed.
Deep learning has since been on a fast-growing trajectory, often providing the best performance.
Hopefully, our work can provide directions on which neural tools are most appropriate and which graph features are most important for the task at hand.
Specifically, we explore whether representing graphs as triplets is more useful than using the full topological information of the graph, and what features can be ignored and still achieve an acceptable disambiguation rate.

Secondly, we create a new dataset to help us in our endeavor.
Among the datasets available for this task we took inspiration from \wikidisamb{} \cite{ferragina_tagme:_2010}.
In that work, \citeauthor{ferragina_tagme:_2010} tackle the problem of cross referencing text fragments with \wikipedia{} pages.
Specifically, they deal with \emph{very short} sentences ($30$-$40$ words).
We build on their work by translating the pointers of \wikipedia{} pages to \wikidata{} items, thereby creating an \emph{ad hoc} dataset based on \wikidisamb{}. 
We call our derivative dataset \wikidatadisamb{}. 
This new dataset creates the perfect playground for us to test various models of \ac{NED} on \wikidata{}.

The paper is organized as follows: 
Sec. \ref{sec:problem_statement} provides a concise description of our task, while in Sec. \ref{sec:methods_and_materials} we describe all our models. 
In Sec. \ref{sec:dataset} we explain how the dataset has been created and in Sec. \ref{sec:training} we detail the training used. 
In Sec. \ref{sec:experimental_results} we summarize the results and discuss the relevance of our models for classifying \wikidata{} entries.
Finally, a review of similar results is presented in Sec. \ref{sec:related_works}, and Sec. \ref{sec:concluding_remarks} concludes and suggests further research directions.


\section{Methodology} 
\label{sec:problem_statement}

All models share three main elements: 
A \emph{graph}, a \emph{text}, and an \emph{entity} in the text to disambiguate. 
The disambiguation task is reduced to a consistency test between the input text and the graph.

The graph is composed by nodes connected with edges. 
The node vectors $\{\mathbf{x}_i\}$ are represented by the centroid of the \glove{} word vectors that make up the nodes: 
For example, a node called "New York" is represented by averaging the word vectors of "New" and "York". 
An edge $\mathbf{e}_{ij}$ connects node $i$ with node $j$. The set of edge vectors $\{\mathbf{e}_{ij}\}$ is computed exactly as for the node vectors, by averaging over the word vectors in each of the edge's labels: 
For example, the vector of "instance of" is the average of the \glove{} vectors "instance" and "of".
The values of the adjacency matrix $\mathbf{A}$ of a graph are set to $1$ in the elements $A_{i,j}$ that are connected by a vertex and $0$ otherwise.

The \emph{text} is described as a sequence of word vectors $\{\mathbf{v}_i\}$, represented using the \glove{} embeddings, while the \emph{item} is used to query the \wikidata{} dataset for the corresponding entry. 
In most models we have an embedding for the input text $\mathbf{y}_\mathrm{text}$ and one for the graph $\mathbf{y}_\mathrm{graph}$.

All our models receive as an input the node vectors $\{\mathbf{x}_i\}$, the word embeddings $\{\mathbf{v}_i\}$ (and possibly the edge vectors $\{\mathbf{e}_{i,j}\}$).
The output of our models is a binary vector, which tells us whether the input graph is consistent with the entity in the text.

\section{Models} 
\label{sec:methods_and_materials}

The size of the training dataset ($2 \times 100000$ items) allows us to experiment with relatively complex models.
In the end we train nine different models, five of which are baselines. The configuration of each model is summed up in \appendixname~\ref{app:configuration}. 

\subsection{Graph related models} 

The \wikidata{} graphs need to be processed by a neural network.
To do so we can either represent the graph as list of triplets - thereby effectively losing the topology of the network - or by employing a method that encodes the topology in the final embeddings. 
In the following we address both representations by employing a \ac{RNN} and a \ac{GCN} \cite{KipfW16} respectively.

In all these models the input text is treated in the same way:
The text word vectors $\{\mathbf{v}_i\}$ are first fed to a \ac{Bi-LSTM}, with outputs $\{\mathbf{y}_i\}$.
These outputs are then weighted by a mask: 
A set of scalars $\{a_i\}$ which are $1$ where the item is supposed to be and $0$ otherwise. 
For example the sentence "The comic book hero Captain Marvel is ..." would have $\{a_i\} = [0, 0, 0, 0, 1, 1, 0, ...]$.
This mask acts as a "manually induced" attention of the item to disambiguate for.
The final output of the \ac{LSTM} mechanism is the the average 
\begin{equation}
\mathbf{y}_\mathrm{text} = N_\mathrm{text}^{-1}\,\sum_{i=0}^{N_\mathrm{text}}\, a_i\, \mathbf{y}_i\,, 
\end{equation}
given a sentence with length $N_\mathrm{text}$.

The following items are our graph-based models:

\begin{itemize}
\item \textbf{Text \ac{LSTM} + \ac{RNN} of triplets}:
\begin{figure}[ht!]
    \centering
        \includegraphics[width=0.4\textwidth]{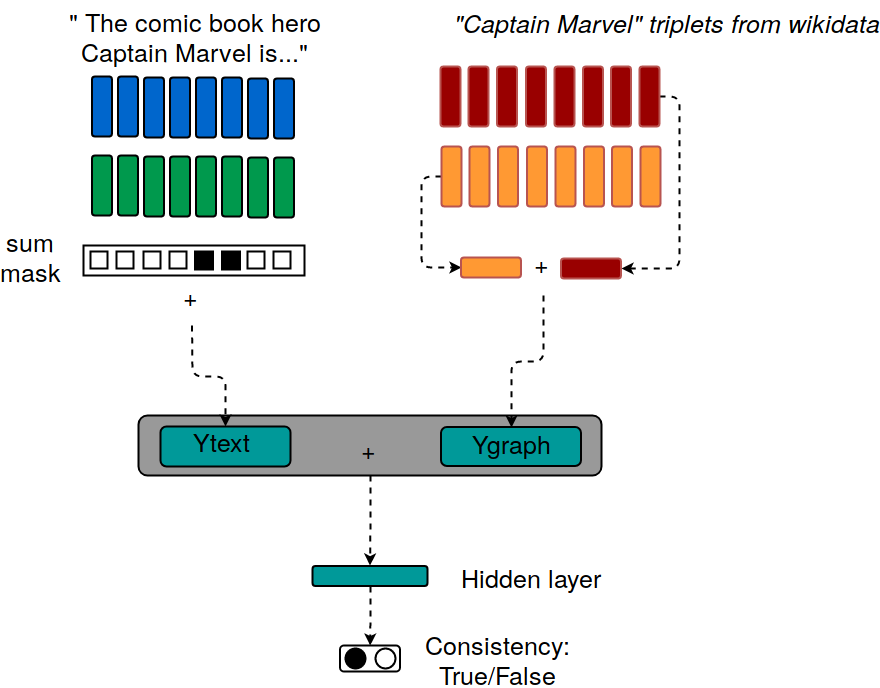}%
    \caption{The \wikidata{} graph can be processed by a \ac{Bi-LSTM} as a list of triplets}
    \label{fig:lstm_rnn}
\end{figure}

In this model, represented in Fig. \ref{fig:lstm_rnn}, a \ac{Bi-LSTM} \cite{augenstein_stance_2016} is applied over the sequence of \emph{triplets} in the graph. 
In the list of input vectors $\mathbf{x}^\mathrm{triplet}$ each item is the concatenation of three elements:
\begin{equation}
\mathbf{x}^\mathrm{triplet}_{i,j} = \mathbf{x}_i
                       \, \oplus \mathbf{e}_{i,j} 
                       \, \oplus \mathbf{x}_j \,
\end{equation}
\noindent where $i, j$ are all the indices between connected nodes in a directed graph. 
The final states of the \ac{Bi-LSTM} are then concatenated and then fed to a dense layer, whose output is the graph embedding $\mathbf{y}_\mathrm{graph}$.

While this model captures the information of single hops in the graph, it is not suited for capturing the topology of the network. 
For example, nodes that are topologically close might appear far away in the set of triplets. 
More importantly, the final embeddings might depend on the specific ordering of the triplets, losing the information about the network shape.

~\\
\item \textbf{Text \ac{LSTM} + \ac{RNN} with attention}:
\begin{figure}[ht!]
    \centering
        \includegraphics[width=0.4\textwidth]{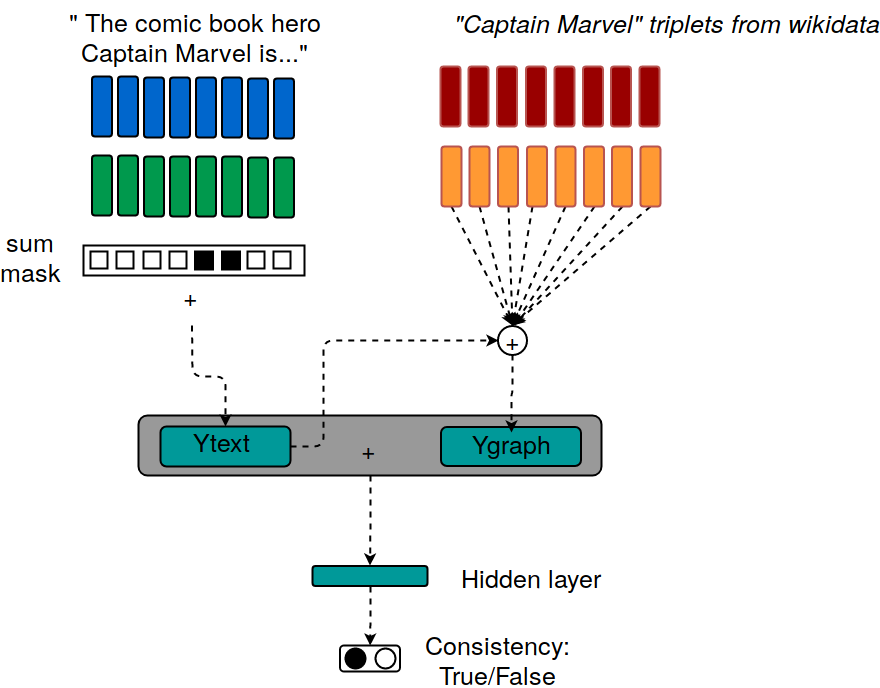}%
    \caption{An additional attention mechanism is added to the \ac{RNN} of triplets model.}
    \label{fig:lstm_rnn_attention}
\end{figure}
We improve upon the prior model by adding an attention mechanism \cite{bahdanau_neural_2014,vaswani_attention_nodate} after the \ac{LSTM} for triplets (Fig. \ref{fig:lstm_rnn_attention}).
The output vectors $z_i$ of the \ac{LSTM} are weighted by an attention coefficient (scalar) $b_i$ and then summed together to create the context vector for the graph.
\begin{equation}
\mathbf{y}_\mathrm{graph} = N_\mathrm{triplets}^{-1}\,\sum_{i=0}^{N_\mathrm{triplets}}\, b_i\, \mathbf{z}_i\,, 
\end{equation}
with
\begin{equation}
\begin{array}{l}
\mathbf{b} = \mathrm{softmax}(\mathbf{c})\, \\
c_i = \mathrm{ReLU}(\mathbf{W}_\mathrm{triplets}\,\mathbf{z}_i + \mathbf{W}_\mathrm{text} \, \mathbf{y}_\mathrm{text} + \mathbf{b}_\mathrm{triplets} )
\,,
\end{array}
\end{equation}
Where the matrices $\mathbf{W}_\mathrm{triplets}$ and $\mathbf{W}_\mathrm{text}$ and the vector $\mathbf{b}_\mathrm{triplets}$ are learned in training. 
We expect this attention method to improve the disambiguation task by giving more weight to relevant triplets.
~\\

\item \textbf{Text \ac{LSTM} + \ac{GCN}}:

\begin{figure}[ht!]
    \centering
        \includegraphics[width=0.4\textwidth]{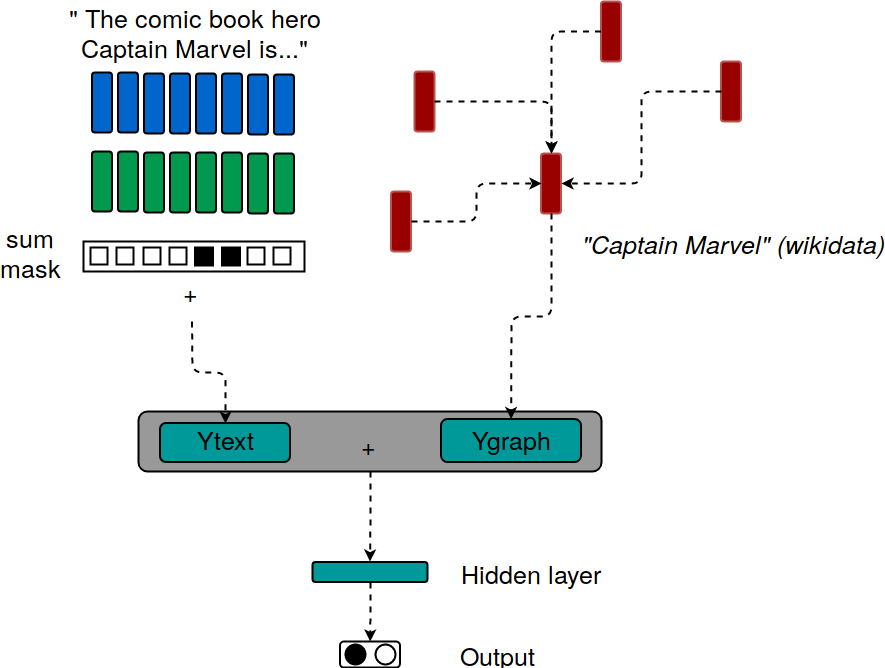}%
    \caption{Diagram of the \ac{GCN} approach to disambiguation.}
    \label{fig:lstm_gcn}
\end{figure}

\begin{figure}[ht!]
    \centering
        \includegraphics[width=0.35\textwidth]{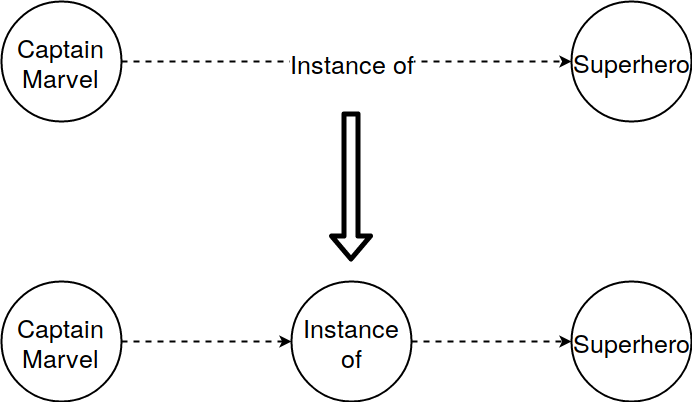}%
    \caption{Reification of the \wikidata{} relations as a pre-processing step for the \ac{GCN}.}
    \label{fig:reify}
\end{figure}

We couple the \ac{Bi-LSTM} with a \ac{GCN} \cite{KipfW16} to compare the sentence to the \wikidata{} graph. 
The base diagram of the network is in Fig. \ref{fig:lstm_gcn}. 

\acp{GCN} have been able to provide state-of-the-art results for Entity Prediction \cite{schlichtkrull_modeling_2017}, Semantic Role Labelling \cite{marcheggiani_encoding_2017}, matrix completion for recommender systems \cite{berg_graph_2017}, and relational inference \cite{kipf_neural_2018}. 
It seems therefore natural to use graph convolutions for our \ac{NED} task. 
Specifically, the convolutions can be employed to create an embedding vector of the relevant \wikidata{} graph.

A graph convolutional network works by stacking convolutional layers based on the topology of the network. Typically, by stacking together $N$ layers the network can propagate the features of nodes that are at most $N$ hops away. The information at the $k^{st}$ layer is propagated to the next one according to the equation
\begin{equation}
    \mathbf{h}_v^{k+1} = \mathrm{ReLU} \left(
        \sum_{u \in \mathcal{N}(v)} \left( 
            \mathbf{W}^{k}\,\mathbf{h}_u^{k} + \mathbf{b}^{k} 
        \right) 
    \right) \,,
    \label{eq:gcn}
\end{equation}

\noindent where $u$ and $v$ are two indices of nodes in the graph. 
$\mathcal{N}$ is the set of nearest neighbors of node $v$, plus the node $v$ itself.
The vector $\mathbf{h}_u^{k}$ represents node $u$'s embeddings at the $k^\mathrm{st}$ layer. The matrix $\mathbf{W}$ and vector $\mathbf{b}$ are learned during training and map the embeddings of node $u$ onto the adjacent nodes in the graph. In this paper the we only consider the outgoing edges from each node.

With the topology of the \wikidata{} graph, the information of each node is propagated onto the central item. 
Ideally, after the graph convolutions, the vector at the position of the central item summarizes the information in the graph.

One the challenges of the original formulation of \ac{GCN} is about including the information contained in the edges' labels (which are not present in Eq. \ref{eq:gcn}).
One way to solve this issue is to see the convolutions as a form of \emph{message passing} \cite{gilmer_neural_2017,kipf_neural_2018} .

We do not explicitly use the \emph{message passing} technique. 
Instead we opt for a similar solution: we reify the relations to appear as additional nodes (see Fig. \ref{fig:reify}).
In this way the edges become nodes themselves. 
The end result is comparable to the message passing model, where information flows from a vertex to an edge and is eventually dispatched to another node.

The original formulation of \ac{GCN} can therefore be applied to this modified graph.

~\\
\acp{GCN} are designed to capture the topology of the graph. The final vector contains information that comes from the node vectors, the edge vectors, and the adjacency matrix. These components end up building the vector embedding $\mathbf{y}_\mathrm{graph}$.

~\\

\item \textbf{Text \ac{LSTM} + \ac{GCN} with attention}:
We aim to improve upon the prior model by adding an attention mechanism after the \ac{GCN} layers.
This can be obtained by adding a set of weights $\alpha_{vu}$ so that Eq. \ref{eq:gcn} becomes
\begin{equation}
    \mathbf{h}_v^{k+1} = \mathrm{ReLU} \left(
        \sum_{u \in \mathcal{N}(v)} 
        \alpha_{vu}
        \left( 
            \mathbf{W}^{k}\,\mathbf{h}_u^{k} + \mathbf{b}^{k} 
        \right) 
    \right) \,.
\end{equation}
~\\
The coefficients $\alpha_{uv}$ signify the attention to be paid to the information being passed from node $u$ to node $v$. 
This attention needs to be a function of the vector and edge nodes, as well as a function of the input text. We choose the following method for \ac{GCN} attention:

\begin{equation}
\begin{array}{l}
\boldsymbol{\alpha} = (\mathbb{1} +\mathbf{A}) \,\odot \,\mathrm{softmax}(\mathbf{E})\, \\
\mathbf{E} = \mathbf{B}^\top\,\mathbf{B}\\
\mathbf{B} = \mathrm{ReLU}(\mathbf{W}_\mathrm{graph}\,\mathbf{H}^k 
						   + \mathbf{W}_\mathrm{text} \, \mathbf{Q}_\mathrm{text} + \mathbf{C}_\mathrm{graph})
\,,
\end{array}
\end{equation}

where the $\mathrm{softmax}$ function acts on the last dimension of $\mathbf{E}$ and $\mathbf{A}$ is the original adjacency matrix for the graph.
The matrix $\mathbf{B}$ models the information propagating from a node in the context of the input text:
The columns of the matrix $\mathbf{H}^k \in \mathbb{R}^{m \times n}$ are the layer vectors $\mathbf{h}_u^k$, with $n$ the number of nodes and $m$ the dimension of the layer embeddings;
$\mathbf{Q} \in \mathbb{R}^{q \times n}$ is a matrix where all the columns are identical and equal to the input text embeddings $\mathbf{y}_\mathrm{text}$, with $q$ the dimension of the text embeddings. In the matrices $\mathbf{W}_\mathrm{graph} \in \mathbb{R}^{d \times m}$ and $\mathbf{W}_\mathrm{text} \in \mathbb{R}^{d \times q}$ and in the bias matrix $\mathbf{C}_\mathrm{graph} \in \mathbb{R}^{d \times n}$ $d$ is an arbitrary intermediate dimension. 

If $\mathbf{B} \in \mathbb{R}^{d \times n}$ models the outgoing information from each node, the matrix $\mathbf{B}^\top$ then models the information arriving to the nodes. 
In the end $\mathbf{B}^\top\,\mathbf{B}$ ($\in \mathbb{R}^{n \times n}$) is the set of weights for messages being propagated among all the nodes.
A final element-wise multiplication with $(\mathbb{1} +\mathbf{A})$ masks out the elements that are not connected.
~\\
To the best of our knowledge, this formulation of \ac{GCN} attention is original. 
\end{itemize}
~\\~\\

\subsection{Baseline models} 

The evaluation of previous models would not be meaningful without a set of baselines. 
To this end, we want to know how much of the graph information is useful to achieve the best accuracy.

\begin{itemize}
\item \textbf{Vector distance baseline}:
This is the simplest method, based on the hypothesis that the input text might be somehow semantically closer to the correct graph than to the wrong one.
We therefore take as inputs the simple average $\bar{\mathbf{v}}$ of the text vectors and the average $\bar{\mathbf{x}}$ of all the nodes' vectors in the graph. 
The classification task is then performed by finding a distance $d$ according to which 
\begin{equation}
||\,\bar{\mathbf{v}} - \bar{\mathbf{x}}\,|| < d
\end{equation}
for the correct graph, and
\begin{equation}
||\,\bar{\mathbf{v}} - \bar{\mathbf{x}}\,|| > d
\end{equation}
for the incorrect graph.
We chose $d=0.945$ as the distance that maximizes the $F1$ score in the training dataset.

\item \textbf{Feedforward of averages}:
We take the average $\bar{\mathbf{v}}$ of the words in the sentence and the average $\bar{\mathbf{x}}$ of all the nodes in the graph, concatenate them, and feed them to a feedforward neural net with one hidden layer.
The final output is binary, meaning that the sentence can be either consistent or inconsistent with the \wikidata{} graph.
\end{itemize}

\begin{itemize}
\item \textbf{Text \ac{LSTM} + Centroid}:
In this model (and in the following ones) the input text is processed by the same \ac{Bi-LSTM} method of Fig. \ref{fig:lstm_rnn}, Fig. \ref{fig:lstm_rnn_attention}, and \ref{fig:lstm_gcn}.
Here the graph information is instead collapsed onto the average vector $\mathbf{y}_\mathrm{graph} = \bar{\mathbf{x}}$ as in the baseline. 

\item \textbf{Text \ac{LSTM} + Linear attention}: 
Instead of using the average $\bar{\mathbf{x}}$ for representing the graph, we employ an attention model over the node vectors. 
The output of this attention model is
\begin{equation}
\mathbf{y}_\mathrm{graph} = N_\mathrm{nodes}^{-1}\,\sum_{i=0}^{N_\mathrm{nodes}}\, b_i\, \mathbf{x}_i\,, 
\end{equation}
with
\begin{equation}
\begin{array}{l}
\mathbf{b} = \mathrm{softmax}(\mathbf{c})\, \\
c_i = \mathrm{ReLU}(\mathbf{W}_\mathrm{nodes}\,\mathbf{x}_i + \mathbf{W}_\mathrm{text} \, \mathbf{y}_\mathrm{text} + \mathbf{b}_\mathrm{nodes} )
\,,
\end{array}
\end{equation}
Where $\mathbf{W}_\mathrm{nodes}$, $\mathbf{W}_\mathrm{text}$, and $\mathbf{b}_\mathrm{nodes}$ are learned through backpropagation. 
This attention technique ideally improves the classification task by giving more weight to relevant nodes.
\,
\,
\item \textbf{Text \ac{LSTM} + \ac{RNN} of nodes}:
Instead of taking the average of the nodes in the graph, we use a \ac{Bi-LSTM} on the nodes and then concatenate the final hidden layers to create the representation vector $\mathbf{y}_\mathrm{graph}$ (same structure as Fig. \ref{fig:lstm_rnn}, but with node vectors in place of triplet vectors).
\end{itemize}

\section{The dataset} 
\label{sec:dataset}

We create a new dataset from the information in the \wikidisamb{} set.
Originally, the dataset addressed the need of a corpus of very short texts (a few 30-40 words) where a specific entity was linked to the correct \wikipedia{} page. The original dataset contains about 2 million entries and presents three elements for each one: 
an English sentence, the name of the entity to disambiguate, and the correct \wikipedia{} item corresponding to  the entity. 

One example is presented in Table \ref{tab:wikidisamb_item}.
\begin{table*}[h!]
    \centering
    \begin{tabular}{ c c c}
        \toprule
            \multicolumn{1}{c}{\textbf{Text}} & \multicolumn{1}{c}{\textbf{Entity}} & \multicolumn{1}{c}{\textbf{Wikipedia ID}}\\
        \midrule
            \shortstack{fantasy novelist David Gemmell. Achilles is \\
            featured heavily in the novel The Firebrand by Marion \\
            Zimmer Bradley. The comic book hero Captain Marvel is \\
            endowed with the courage of Achilles, as well}
            &
            Captain Marvel
            &
            403585 \\
        \bottomrule
    \end{tabular}
    \caption{Example item from the original dataset \wikidisamb{}.}
    \label{tab:wikidisamb_item}
\end{table*}
\begin{table*}[h!]
    \centering
    \begin{tabular}{c c c c}
        \toprule
            \multicolumn{1}{c}{\textbf{Text}} & \multicolumn{1}{c}{\textbf{Entity}} & \multicolumn{1}{c}{\textbf{Correct \wikidata{} ID}}& \multicolumn{1}{c}{\textbf{Wrong \wikidata{} ID}}\\
        \midrule
            \shortstack{fantasy novelist David Gemmell. Achilles is \\
            featured heavily in the novel The Firebrand by Marion \\
            Zimmer Bradley. The comic book hero Captain Marvel is \\
            endowed with the courage of Achilles, as well}
            &
            Captain Marvel
            &
            Q534153 
            &
            Q41421
            \\
        \bottomrule
    \end{tabular}
    \caption{Example item from our dataset \wikidatadisamb{}. The original \wikipedia{} entry is translated to a \wikidata{} ID and additional wrong \wikidata{} item with the same name (or alias) is added.}
    \label{tab:wikdatadisamb_item}
\end{table*}
\,%
\,%
The ambiguous item here is the name \emph{Captain Marvel}, which is a character from Marvel comics and -- for example -- a nickname for \emph{Michael Jordan}, the basketball player. 
The correct interpretation in the example sentence is the former.

Our dataset provides a conversion from the \wikipedia{} page to a \wikidata{} item, when this conversion exists. 
If the conversion is not possible the original entry is simply skipped.
In order to have a consistent disambiguation task we also select an incorrect \wikidata{} item to pair with the correct one, linked to it by having the same name (or same alias).

One example item is as in Table \ref{tab:wikdatadisamb_item}: 
The correct item is \textbf{Q534153}, the \wikidata{} entry of \emph{Captain Marvel}. 
The incorrect entry is \textbf{Q41421}, which is the entry for \emph{Michael Jordan} - also know as \emph{Captain Marvel}.
This incorrect entry is selected to not be trivial, i.e., a disambiguation page or an entry with no triplets.
In this way we obtain a balanced dataset, where the correct entity appears as many times as the wrong one. 

After applying those selection constraints - and keeping in mind scalability issues - we chose 120000 items, of which 100 thousand in the training set, and 10000 entries each for the development and test sets. 
This information is then fed to our models.

Each model performance is measured on how well it can predict if a \wikidata{} graph is consistent with the entity in the input sentence. 
Since we measure the consistency with \emph{correct} and \emph{wrong} \wikidata{} IDs separately, the size of the training set effectively doubles, with 200 thousand graphs to compare with their respective sentences. Likewise, in the development and test sets we compare the consistency predictions of 20 thousand graphs with the relevant texts.

\section{Training} 
\label{sec:training}
We use \tensorflow{}~\cite{tensorflow2015-whitepaper} to implement our neural network.
The weights are initialized randomly from the uniform distribution and the initial state of the \acp{LSTM} are set to zero. 
We use binary cross entropy as the final loss function.
An Adam optimizer \cite{KingmaB14} is used with a step of $10^{-4}$. 
Whenever applicable, we employ a batch size of 10. 

The dimension of the \glove{} vectors used in our experiments is 300, and in all our tests we cut the \wikidata{} graph after 2 hops from the central node.
This hopping distance has been selected to maximize the perfomance of the \ac{GCN} based models.



\section{Experimental Results} 
\label{sec:experimental_results}

\label{sec:disambiguation}
\begin{table*}[ht!]
    \centering
    \begin{tabular}{c c c c c c c}
        \toprule
              & \multicolumn{3}{c}{\textsc{\textbf{Dev}}} & \multicolumn{3}{c}{\textsc{\textbf{Test}}} \\
            \multicolumn{1}{c}{\textbf{Description}} & \textbf{prec} & \textbf{rec} & \textbf{$F_1$} & \textbf{prec} & \textbf{rec} & \textbf{$F_1$} \\	
        \midrule
        	\emph{Vector distance baseline}& $57.4$ & $53.4$ & $55.3$ & $57.4$ & $53.4$ & $55.4$ \\
        	\emph{Feedforwad of averages} & $82.9$ & $86.9$ & $84.8$ & $82.2$ & $87.1$ & $84.5$ \\
        	\emph{Text \ac{LSTM}}+\emph{Centroid} & $87.5$ & $91.4$ & $89.5$ & $87.3$ & $91.8$ & $89.5$ \\
            \emph{Text \ac{LSTM}}+\emph{Linear attention} & $80.4$ & $91.7$ & $85.7$ & $79.7$ & $90.6$ & $84.6$ \\
        	\emph{Text \ac{LSTM}}+\emph{\ac{RNN} of nodes} & $80.4$ & $89.6$ & $84.7$ & $79.6$ & $89.5$ & $84.2$ \\
        	\emph{Text \ac{LSTM}}+\emph{\ac{RNN} of triplets} & $\mathbf{90.7}$ & $92.2$ & $91.4$ & $90.1$ & $92.0$ & $91.1$ \\
            \emph{Text \ac{LSTM}}+\emph{\ac{RNN} of triplets with attention} & $90.2$ & $\mathbf{93.1}$ & $\mathbf{91.6}$ & $\mathbf{90.2}$ & $\mathbf{93.0}$ & $\mathbf{91.6}$ \\
            \emph{Text \ac{LSTM}}+\ac{GCN} & $71.0$ & $87.6$ & $78.4$ & $70.0$ & $87.8$ & $77.8$ \\
            \emph{Text \ac{LSTM}}+\ac{GCN} with attention & $73.7$ & $91.1$ & $81.5$ & $74.8$ & $88.2$ & $81.0$ \\
        \bottomrule
    \end{tabular}
    \caption{Results of our architectures expressed as a percentage (best results in bold).}
    \label{tab:results} 
\end{table*}

The results of our experiments are presented in Table \ref{tab:results}. We took two evaluations for each model and show the average result. 
The difference between the lowest and highest score varies between $0.4 \%$ and $0.8 \%$ for the different models.
In absence of more complete statistics, we choose the middle value $0.6 \%$ as an estimate for the statistical error to attribute to all our measurements. All results are approximated to the first significant digit of the error.

The simple \emph{vector distance baseline} is seen here performing narrowly better than random chance, with $F1 = 55.4\%$ on the test set. 
This is not unexpected since the model only captures the distance between two centroids.

The \emph{feedforward of average}s works much better, with $F1 = 84.5\%$ on the test set. 
Given how little information is fed as an input, the result has more to do with the quality of the \glove{} vectors than our model.

The other models are more complex, and this additional complexity seem to have non-trivial consequences in the results. 
For example, the \emph{text \ac{LSTM} + Centroid} model scores an $F1 = 89.5\%$ on the test set, the third highest in this paper. 
This is an increase of $5\%$ over the simple \emph{feedforward} model, meaning that the text \ac{LSTM} part (the only change from the prior model) is extremely relevant in processing the input text.

The following two models, \emph{linear attention} and \emph{\ac{RNN} of nodes}, do not provide any significant improvement upon the \emph{feedforward of averages}. 
The modest results of the \emph{linear attention} model are particularly interesting, suggesting that the classification task does not seem to rely on specific easy-to-identify nodes, and that the whole node set information seem to play a role for an accurate result.

The second best results of the paper is given by the \emph{\ac{RNN} of triplets} model, with $F1 = 91.1 \%$ on the test set. This model uses the whole graph information taking as an input the set of triplets that compose the \wikidata{} graph. 
The \emph{\ac{RNN} of triplets with attention} seems to perform even better, reaching $F1 = 91.6 \%$. 
A straightforward conclusion is that the classification task is mildly helped by paying attention on specific triplets.
In retrospect this is unsurprising, as relations like \emph{instance\_of}, \emph{subclass\_of} give a relevant hint to the type of entity described by the graph.
The improvement is however small, and it seems to indicate that - on average - there is no "critical triplet" for \ac{NED}.

Conversely, the \emph{Text \ac{LSTM}}+\ac{GCN} model performs poorly, with $F1 = 77.8\%$. 
The reason of this drop in performance is complex, and we believe it rests in the way \acp{GCN} create the final embedding vector.
The \ac{GCN} embeddings sum up information that comes from the graph nodes, edges, and topology of the network. 
This last piece of information is not considered in the triplet model, and we believe it is what confuses the graph convolutional model:
For example, looking at Fig. \ref{fig:graphs} the \ac{GCN} might decide that the main difference between the two graphs is in the network topology, not in the content of the nodes;
conversely graphs with similar topology can be considered closer to each other than graphs with similar triplets.
In short, in our experiments the \ac{GCN} seems to give too much importance to the shape of the graphs in training, ending up being confused when testing.
It seems that the graph convolutional model would perform better in a dataset where the topology of the graphs plays a more relevant role. 
In our datasets the graphs are simple trees, and the key pieces of information seem to come from relation triplets.

The \emph{Text \ac{LSTM}+\ac{GCN} with attention} model seems to perform about $3 \%$ better. 
The attention model effectively adapts the topology of the network to the input text, alleviating some of the issues with the prior model. 
Even so, the attentive \ac{GCN} does not perform well. 
We would be excited to apply a \ac{GCN}-based model to a dataset with a richer network topology in a future work.




\section{Related Works} 
\label{sec:related_works}
Disambiguating among similar entities is a crucial feature in extracting relevant information from texts:
Ambiguous information needs to be resolved, requiring additional steps that go beyond grammatical parsing.
Correctly sieving information from huge corpora of text is especially suited for a Deep learning approach, given the data-hungry nature of neural networks.

\ac{EL} is however not a recent endeavor. The work of \citeauthor{bunescu_using_nodate} \cite{bunescu_using_nodate} introduced the idea of disambiguating text through the use of knowledge bases.
An entity in a sentence is compared to entries in a corpus and the correct meaning is resolved by using an appropriate similarity function.

Shortly thereafter, the authors of \cite{milne_learning_2008} introduced the idea of learning to link entities on \wikipedia{}. In this spirit, the \wikidisamb{} dataset \cite{ferragina_tagme:_2010} was created and further used in the TAC2011 Knowledge Base Population Track \cite{mcnamee_overview_2009}.

The previous works use a corpus of unstructured texts (\wikipedia{}) to disambiguate entities in sentences. 
However, there is a long tradition of using structured data for entity classification and linking. 
One example is \cite{arenas_tabel:_2015}, where they address the problem of entity linking within \emph{web tables}. Their problem is similar to the the one we tackle in this paper, although restricted to the types of tables one can find in \wikipedia{} pages or more general \emph{html} pages.
The \wikipedia{} social network is exploited in \cite{geis_little_2016}, where they study the linking of person entities and disambiguation of homonyms.
A work closely related to ours is \cite{mika_agdistis_2014}, in which the authors employ a search-based algorithm for the disambiguation task.
We refrained however from using the datasets in that paper because the amount of annotated data used therein seemed to be insufficient for training a neural net.

In \cite{schuhmacher_knowledge-based_2014} the authors present a novel way to build a semantic graph to represent a document content.
Those graphs are then used to rank related entities, as well as providing a document similarity score.
The ranking algorithm is built around the idea that similar entities are close to each other in the semantic graph (hopping distance). 
A similar work is presented in \cite{ren_cotype:_2017}.
The authors created a system (CoType) where relations and entities are extracted together by means of comparing the entities in a sentence to the items in a knowledge graph. 
Furthermore, the task of using semantic graphs to mine topics is explored in \cite{chen_semantic_2017}.

The role of deep learning in \ac{EL} is also studied in some recent works. The authors \citeauthor {globerson_collective_2016} invent a novel attention mechanism for entity resolution. 
In \cite{raiman_deeptype:_2018} a system of entities that are easy to learn is created, and eventually they are able to improve upon the state-of-the-art in the \wikidisamb{} dataset.



The work of \citeauthor{meij2014} minutely reviews the role of \ac{EL} in a semantic search context, arguing that \ac{NED} and Entity Retrieval enables modern search engines to organize their wealth of information around entities.

Most \ac{EL} datasets are based on relatively well-behaved text. 
The challenge of \ac{NED} on noisy text is addressed in \cite{eshel_named_2017}, where they present a new dataset with more realistic \emph{html} page fragments. 
Another problem is disambiguating entities in a question answering system. This issue is studied in \cite{klang_named_nodate}.

\subsection{Comparison with \wikidisamb{} results}
\label{sec:comparison}
The dataset we present in this paper is derived from the \wikidisamb{} corpus.
A comparison with prior results evaluated on the original dataset seems due, albeit somewhat contrived:
In the original dataset the context for disambiguation comes from \wikipedia{} pages, whereas in our work we build an embedding vector from \wikidata{} graphs.

In \cite{raiman_deeptype:_2018} the authors summarize recent \ac{NED} results running  the \wikidisamb{} dataset using the algorithms of the original papers. 
They report $F1 = 84.6 \%$ for \cite{milne_learning_2008} and $F1 = 90.9$ for \cite{ferragina_tagme:_2010}. 
The state-of-the-art still lies in the work of \citeauthor{raiman_deeptype:_2018}, where they achieve $F1 = 92.4 \%$.

\section{Concluding Remarks} 
\label{sec:concluding_remarks}

We have shown that it is possible to disambiguate entities in short sentences by looking at the corresponding entries in \wikidata{}. 
In order to achieve this result, we created a new dataset \wikidatadisamb{}, where we present an equal number of correct and incorrect entity linking candidates.

Our \emph{\ac{RNN} of triplets with attention} model allows us to achieve the best result $F1 = 91.6\%$ over the test set.
This is an improvement from the baseline of the simple \emph{feedforward of averages} model of about $7.1\%$, where the edges of the \wikidata{} graph are not used.

The main contribution of this improvement seems to come from processing the input text with a \ac{Bi-LSTM}. 
Various methods of dealing with the \wikidata{} graph do not seem to correlate with a big improvement in the results. 
Indeed, most baseline models - that only consider the nodes of the graph - seem to perform roughly equally.
The second biggest improvement happens when including information about the relation type with the \emph{\ac{RNN} of triplets}.

The \ac{GCN} based approaches are seen to perform poorly. 
In our dataset the topology of the graphs seems to play a secondary role (most of the graphs in the dataset are simple trees), and the performance of the graph convolutional models drops as a result.
More interesting graph topologies should make the \ac{GCN} perform better.
We aim to address this issue with different datasets in following works.

In the future, we aim to use similar techniques to pursue disambiguation tasks outside the dataset we created.
Moreover, a similar approach can be used to create \emph{graph embeddings} of \wikidata{} items, which can be used for improving semantic search tasks \cite{meij2014}.
In addition, cross-language entity linking could be addressed with techniques similar to ours, by using datasets in different languages \cite{Pappu:2017:LME:3018661.3018724}.
Our aim is to address these challenges in following works.

\section*{Acknowledgements} 
This work was partially supported by InnovateUK grant Ref. 103677.

\bibliographystyle{ACM-Reference-Format}


\begin{thebibliography}{35}


\ifx \showCODEN    \undefined \def \showCODEN     #1{\unskip}     \fi
\ifx \showDOI      \undefined \def \showDOI       #1{#1}\fi
\ifx \showISBNx    \undefined \def \showISBNx     #1{\unskip}     \fi
\ifx \showISBNxiii \undefined \def \showISBNxiii  #1{\unskip}     \fi
\ifx \showISSN     \undefined \def \showISSN      #1{\unskip}     \fi
\ifx \showLCCN     \undefined \def \showLCCN      #1{\unskip}     \fi
\ifx \shownote     \undefined \def \shownote      #1{#1}          \fi
\ifx \showarticletitle \undefined \def \showarticletitle #1{#1}   \fi
\ifx \showURL      \undefined \def \showURL       {\relax}        \fi
\providecommand\bibfield[2]{#2}
\providecommand\bibinfo[2]{#2}
\providecommand\natexlab[1]{#1}
\providecommand\showeprint[2][]{arXiv:#2}

\bibitem[\protect\citeauthoryear{??}{vas}{2017}]%
        {vaswani_attention_nodate}
 \bibinfo{year}{2017}\natexlab{}.
\newblock \showarticletitle{Attention is {All} you {Need}}.
\newblock  (\bibinfo{year}{2017}).
\newblock


\bibitem[\protect\citeauthoryear{Abadi, Agarwal, Barham, Brevdo, Chen, Citro,
  Corrado, Davis, Dean, Devin, Ghemawat, Goodfellow, Harp, Irving, Isard, Jia,
  Jozefowicz, Kaiser, Kudlur, Levenberg, Man\'{e}, Monga, Moore, Murray, Olah,
  Schuster, Shlens, Steiner, Sutskever, Talwar, Tucker, Vanhoucke, Vasudevan,
  Vi\'{e}gas, Vinyals, Warden, Wattenberg, Wicke, Yu, and Zheng}{Abadi
  et~al\mbox{.}}{2015}]%
        {tensorflow2015-whitepaper}
\bibfield{author}{\bibinfo{person}{Mart\'{\i}n Abadi}, \bibinfo{person}{Ashish
  Agarwal}, \bibinfo{person}{Paul Barham}, \bibinfo{person}{Eugene Brevdo},
  \bibinfo{person}{Zhifeng Chen}, \bibinfo{person}{Craig Citro},
  \bibinfo{person}{Greg~S. Corrado}, \bibinfo{person}{Andy Davis},
  \bibinfo{person}{Jeffrey Dean}, \bibinfo{person}{Matthieu Devin},
  \bibinfo{person}{Sanjay Ghemawat}, \bibinfo{person}{Ian Goodfellow},
  \bibinfo{person}{Andrew Harp}, \bibinfo{person}{Geoffrey Irving},
  \bibinfo{person}{Michael Isard}, \bibinfo{person}{Yangqing Jia},
  \bibinfo{person}{Rafal Jozefowicz}, \bibinfo{person}{Lukasz Kaiser},
  \bibinfo{person}{Manjunath Kudlur}, \bibinfo{person}{Josh Levenberg},
  \bibinfo{person}{Dan Man\'{e}}, \bibinfo{person}{Rajat Monga},
  \bibinfo{person}{Sherry Moore}, \bibinfo{person}{Derek Murray},
  \bibinfo{person}{Chris Olah}, \bibinfo{person}{Mike Schuster},
  \bibinfo{person}{Jonathon Shlens}, \bibinfo{person}{Benoit Steiner},
  \bibinfo{person}{Ilya Sutskever}, \bibinfo{person}{Kunal Talwar},
  \bibinfo{person}{Paul Tucker}, \bibinfo{person}{Vincent Vanhoucke},
  \bibinfo{person}{Vijay Vasudevan}, \bibinfo{person}{Fernanda Vi\'{e}gas},
  \bibinfo{person}{Oriol Vinyals}, \bibinfo{person}{Pete Warden},
  \bibinfo{person}{Martin Wattenberg}, \bibinfo{person}{Martin Wicke},
  \bibinfo{person}{Yuan Yu}, {and} \bibinfo{person}{Xiaoqiang Zheng}.}
  \bibinfo{year}{2015}\natexlab{}.
\newblock \bibinfo{title}{{TensorFlow}: Large-Scale Machine Learning on
  Heterogeneous Systems}.
\newblock
\newblock
\urldef\tempurl%
\url{http://tensorflow.org/}
\showURL{%
\tempurl}
\newblock
\shownote{Software available from tensorflow.org.}


\bibitem[\protect\citeauthoryear{Artiles, Amigó, and Gonzalo}{Artiles
  et~al\mbox{.}}{2009}]%
        {artiles_role_2009}
\bibfield{author}{\bibinfo{person}{Javier Artiles}, \bibinfo{person}{Enrique
  Amigó}, {and} \bibinfo{person}{Julio Gonzalo}.}
  \bibinfo{year}{2009}\natexlab{}.
\newblock \showarticletitle{The role of named entities in web people search}.
  In \bibinfo{booktitle}{\emph{Proceedings of the 2009 {Conference} on
  {Empirical} {Methods} in {Natural} {Language} {Processing} {Volume} 2 -
  {EMNLP} '09}}, Vol.~\bibinfo{volume}{2}. \bibinfo{publisher}{Association for
  Computational Linguistics}, \bibinfo{address}{Singapore},
  \bibinfo{pages}{534}.
\newblock
\showISBNx{978-1-932432-62-6}
\urldef\tempurl%
\url{https://doi.org/10.3115/1699571.1699582}
\showDOI{\tempurl}


\bibitem[\protect\citeauthoryear{Augenstein, Rocktäschel, Vlachos, and
  Bontcheva}{Augenstein et~al\mbox{.}}{2016}]%
        {augenstein_stance_2016}
\bibfield{author}{\bibinfo{person}{Isabelle Augenstein}, \bibinfo{person}{Tim
  Rocktäschel}, \bibinfo{person}{Andreas Vlachos}, {and}
  \bibinfo{person}{Kalina Bontcheva}.} \bibinfo{year}{2016}\natexlab{}.
\newblock \showarticletitle{Stance {Detection} with {Bidirectional}
  {Conditional} {Encoding}}. In \bibinfo{booktitle}{\emph{Proceedings of the
  2016 {Conference} on {Empirical} {Methods} in {Natural} {Language}
  {Processing}}}. \bibinfo{publisher}{Association for Computational
  Linguistics}, \bibinfo{address}{Austin, Texas}, \bibinfo{pages}{876--885}.
\newblock
\urldef\tempurl%
\url{https://aclweb.org/anthology/D16-1084}
\showURL{%
\tempurl}


\bibitem[\protect\citeauthoryear{Bahdanau, Cho, and Bengio}{Bahdanau
  et~al\mbox{.}}{2014}]%
        {bahdanau_neural_2014}
\bibfield{author}{\bibinfo{person}{Dzmitry Bahdanau},
  \bibinfo{person}{Kyunghyun Cho}, {and} \bibinfo{person}{Yoshua Bengio}.}
  \bibinfo{year}{2014}\natexlab{}.
\newblock \showarticletitle{Neural {Machine} {Translation} by {Jointly}
  {Learning} to {Align} and {Translate}}.
\newblock \bibinfo{journal}{\emph{arXiv:1409.0473 [cs, stat]}}
  (\bibinfo{date}{Sept.} \bibinfo{year}{2014}).
\newblock
\urldef\tempurl%
\url{http://arxiv.org/abs/1409.0473}
\showURL{%
\tempurl}
\newblock
\shownote{arXiv: 1409.0473.}


\bibitem[\protect\citeauthoryear{Berg, Kipf, and Welling}{Berg
  et~al\mbox{.}}{2017}]%
        {berg_graph_2017}
\bibfield{author}{\bibinfo{person}{Rianne van~den Berg},
  \bibinfo{person}{Thomas~N. Kipf}, {and} \bibinfo{person}{Max Welling}.}
  \bibinfo{year}{2017}\natexlab{}.
\newblock \showarticletitle{Graph {Convolutional} {Matrix} {Completion}}.
\newblock \bibinfo{journal}{\emph{arXiv:1706.02263 [cs, stat]}}
  (\bibinfo{date}{June} \bibinfo{year}{2017}).
\newblock
\urldef\tempurl%
\url{http://arxiv.org/abs/1706.02263}
\showURL{%
\tempurl}
\newblock
\shownote{arXiv: 1706.02263.}


\bibitem[\protect\citeauthoryear{Bhagavatula, Noraset, and Downey}{Bhagavatula
  et~al\mbox{.}}{2015}]%
        {arenas_tabel:_2015}
\bibfield{author}{\bibinfo{person}{Chandra~Sekhar Bhagavatula},
  \bibinfo{person}{Thanapon Noraset}, {and} \bibinfo{person}{Doug Downey}.}
  \bibinfo{year}{2015}\natexlab{}.
\newblock \showarticletitle{{TabEL}: {Entity} {Linking} in {Web} {Tables}}.
\newblock In \bibinfo{booktitle}{\emph{The {Semantic} {Web} - {ISWC} 2015}},
  \bibfield{editor}{\bibinfo{person}{Marcelo Arenas}, \bibinfo{person}{Oscar
  Corcho}, \bibinfo{person}{Elena Simperl}, \bibinfo{person}{Markus
  Strohmaier}, \bibinfo{person}{Mathieu d'Aquin}, \bibinfo{person}{Kavitha
  Srinivas}, \bibinfo{person}{Paul Groth}, \bibinfo{person}{Michel Dumontier},
  \bibinfo{person}{Jeff Heflin}, \bibinfo{person}{Krishnaprasad Thirunarayan},
  \bibinfo{person}{Krishnaprasad Thirunarayan}, {and} \bibinfo{person}{Steffen
  Staab}} (Eds.). Vol.~\bibinfo{volume}{9366}. \bibinfo{publisher}{Springer
  International Publishing}, \bibinfo{address}{Cham},
  \bibinfo{pages}{425--441}.
\newblock
\showISBNx{978-3-319-25006-9 978-3-319-25007-6}
\urldef\tempurl%
\url{https://doi.org/10.1007/978-3-319-25007-6_25}
\showDOI{\tempurl}


\bibitem[\protect\citeauthoryear{Blanco, Ottaviano, and Meij}{Blanco
  et~al\mbox{.}}{2015}]%
        {blanco_fast_2015}
\bibfield{author}{\bibinfo{person}{Roi Blanco}, \bibinfo{person}{Giuseppe
  Ottaviano}, {and} \bibinfo{person}{Edgar Meij}.}
  \bibinfo{year}{2015}\natexlab{}.
\newblock \showarticletitle{Fast and {Space}-{Efficient} {Entity} {Linking} for
  {Queries}}. In \bibinfo{booktitle}{\emph{Proceedings of the {Eighth} {ACM}
  {International} {Conference} on {Web} {Search} and {Data} {Mining} - {WSDM}
  '15}}. \bibinfo{publisher}{ACM Press}, \bibinfo{address}{Shanghai, China},
  \bibinfo{pages}{179--188}.
\newblock
\showISBNx{978-1-4503-3317-7}
\urldef\tempurl%
\url{https://doi.org/10.1145/2684822.2685317}
\showDOI{\tempurl}


\bibitem[\protect\citeauthoryear{Bunescu and Pasca}{Bunescu and Pasca}{2006}]%
        {bunescu_using_nodate}
\bibfield{author}{\bibinfo{person}{Razvan Bunescu} {and}
  \bibinfo{person}{Marius Pasca}.} \bibinfo{year}{2006}\natexlab{}.
\newblock \showarticletitle{Using {Encyclopedic} {Knowledge} for {Named}
  {Entity} {Disambiguation}}.
\newblock  (\bibinfo{year}{2006}), \bibinfo{pages}{8}.
\newblock


\bibitem[\protect\citeauthoryear{Chang, Spitkovsky, Manning, and Agirre}{Chang
  et~al\mbox{.}}{2016}]%
        {chang_comparison_nodate}
\bibfield{author}{\bibinfo{person}{Angel~X Chang}, \bibinfo{person}{Valentin~I
  Spitkovsky}, \bibinfo{person}{Christopher~D Manning}, {and}
  \bibinfo{person}{Eneko Agirre}.} \bibinfo{year}{2016}\natexlab{}.
\newblock \showarticletitle{A comparison of {Named}-{Entity} {Disambiguation}
  and {Word} {Sense} {Disambiguation}}.
\newblock  (\bibinfo{year}{2016}), \bibinfo{pages}{8}.
\newblock


\bibitem[\protect\citeauthoryear{Chen, Jose, Yu, and Yuan}{Chen
  et~al\mbox{.}}{2017}]%
        {chen_semantic_2017}
\bibfield{author}{\bibinfo{person}{Long Chen}, \bibinfo{person}{Joemon~M.
  Jose}, \bibinfo{person}{Haitao Yu}, {and} \bibinfo{person}{Fajie Yuan}.}
  \bibinfo{year}{2017}\natexlab{}.
\newblock \showarticletitle{A {Semantic} {Graph}-{Based} {Approach} for
  {Mining} {Common} {Topics} from {Multiple} {Asynchronous} {Text} {Streams}}.
  \bibinfo{publisher}{ACM Press}, \bibinfo{pages}{1201--1209}.
\newblock
\showISBNx{978-1-4503-4913-0}
\urldef\tempurl%
\url{https://doi.org/10.1145/3038912.3052630}
\showDOI{\tempurl}


\bibitem[\protect\citeauthoryear{Cucerzan}{Cucerzan}{2007}]%
        {cucerzan_large-scale_nodate}
\bibfield{author}{\bibinfo{person}{Silviu Cucerzan}.}
  \bibinfo{year}{2007}\natexlab{}.
\newblock \showarticletitle{Large-{Scale} {Named} {Entity} {Disambiguation}
  {Based} on {Wikipedia} {Data}}.
\newblock  (\bibinfo{year}{2007}), \bibinfo{pages}{9}.
\newblock


\bibitem[\protect\citeauthoryear{Dietz, Kotov, and Meij}{Dietz
  et~al\mbox{.}}{2017}]%
        {Dietz:2017:UKG:3018661.3022756}
\bibfield{author}{\bibinfo{person}{Laura Dietz}, \bibinfo{person}{Alexander
  Kotov}, {and} \bibinfo{person}{Edgar Meij}.} \bibinfo{year}{2017}\natexlab{}.
\newblock \showarticletitle{Utilizing Knowledge Graphs in Text-centric
  Information Retrieval}. In \bibinfo{booktitle}{\emph{Proceedings of the Tenth
  ACM International Conference on Web Search and Data Mining}}
  \emph{(\bibinfo{series}{WSDM '17})}. \bibinfo{publisher}{ACM},
  \bibinfo{address}{New York, NY, USA}, \bibinfo{pages}{815--816}.
\newblock
\showISBNx{978-1-4503-4675-7}
\urldef\tempurl%
\url{https://doi.org/10.1145/3018661.3022756}
\showDOI{\tempurl}


\bibitem[\protect\citeauthoryear{Dorssers, de~Vries, and Alink}{Dorssers
  et~al\mbox{.}}{2017}]%
        {dorssers_ranking_nodate}
\bibfield{author}{\bibinfo{person}{Frank Dorssers}, \bibinfo{person}{Arjen~P de
  Vries}, {and} \bibinfo{person}{Wouter Alink}.}
  \bibinfo{year}{2017}\natexlab{}.
\newblock \showarticletitle{Ranking {Triples} using {Entity} {Links} in a
  {Large} {Web} {Crawl}}. In \bibinfo{booktitle}{\emph{Triple Scorer task at
  {WSDM} Cup 2017}}. \bibinfo{pages}{5}.
\newblock


\bibitem[\protect\citeauthoryear{Eshel, Cohen, Radinsky, Markovitch, Yamada,
  and Levy}{Eshel et~al\mbox{.}}{2017}]%
        {eshel_named_2017}
\bibfield{author}{\bibinfo{person}{Yotam Eshel}, \bibinfo{person}{Noam Cohen},
  \bibinfo{person}{Kira Radinsky}, \bibinfo{person}{Shaul Markovitch},
  \bibinfo{person}{Ikuya Yamada}, {and} \bibinfo{person}{Omer Levy}.}
  \bibinfo{year}{2017}\natexlab{}.
\newblock \showarticletitle{Named {Entity} {Disambiguation} for {Noisy}
  {Text}}.
\newblock \bibinfo{journal}{\emph{arXiv:1706.09147 [cs]}} (\bibinfo{date}{June}
  \bibinfo{year}{2017}).
\newblock
\urldef\tempurl%
\url{http://arxiv.org/abs/1706.09147}
\showURL{%
\tempurl}
\newblock
\shownote{arXiv: 1706.09147.}


\bibitem[\protect\citeauthoryear{Ferragina and Scaiella}{Ferragina and
  Scaiella}{2010}]%
        {ferragina_tagme:_2010}
\bibfield{author}{\bibinfo{person}{Paolo Ferragina} {and} \bibinfo{person}{Ugo
  Scaiella}.} \bibinfo{year}{2010}\natexlab{}.
\newblock \showarticletitle{{TAGME}: on-the-fly annotation of short text
  fragments (by wikipedia entities)}. \bibinfo{publisher}{ACM Press},
  \bibinfo{pages}{1625}.
\newblock
\showISBNx{978-1-4503-0099-5}
\urldef\tempurl%
\url{https://doi.org/10.1145/1871437.1871689}
\showDOI{\tempurl}


\bibitem[\protect\citeauthoryear{Geiß and Gertz}{Geiß and Gertz}{2016}]%
        {geis_little_2016}
\bibfield{author}{\bibinfo{person}{Johanna Geiß} {and}
  \bibinfo{person}{Michael Gertz}.} \bibinfo{year}{2016}\natexlab{}.
\newblock \showarticletitle{With a {Little} {Help} from my {Neighbors}:
  {Person} {Name} {Linking} {Using} the {Wikipedia} {Social} {Network}}.
  \bibinfo{publisher}{ACM Press}, \bibinfo{pages}{985--990}.
\newblock
\showISBNx{978-1-4503-4144-8}
\urldef\tempurl%
\url{https://doi.org/10.1145/2872518.2891109}
\showDOI{\tempurl}


\bibitem[\protect\citeauthoryear{Gilmer, Schoenholz, Riley, Vinyals, and
  Dahl}{Gilmer et~al\mbox{.}}{2017}]%
        {gilmer_neural_2017}
\bibfield{author}{\bibinfo{person}{Justin Gilmer}, \bibinfo{person}{Samuel~S.
  Schoenholz}, \bibinfo{person}{Patrick~F. Riley}, \bibinfo{person}{Oriol
  Vinyals}, {and} \bibinfo{person}{George~E. Dahl}.}
  \bibinfo{year}{2017}\natexlab{}.
\newblock \showarticletitle{Neural Message Passing for Quantum Chemistry}.
\newblock  (\bibinfo{year}{2017}).
\newblock
\showeprint[arxiv]{1704.01212}
\urldef\tempurl%
\url{http://arxiv.org/abs/1704.01212}
\showURL{%
\tempurl}


\bibitem[\protect\citeauthoryear{Globerson, Lazic, Chakrabarti, Subramanya,
  Ringaard, and Pereira}{Globerson et~al\mbox{.}}{2016}]%
        {globerson_collective_2016}
\bibfield{author}{\bibinfo{person}{Amir Globerson}, \bibinfo{person}{Nevena
  Lazic}, \bibinfo{person}{Soumen Chakrabarti}, \bibinfo{person}{Amarnag
  Subramanya}, \bibinfo{person}{Michael Ringaard}, {and}
  \bibinfo{person}{Fernando Pereira}.} \bibinfo{year}{2016}\natexlab{}.
\newblock \showarticletitle{Collective {Entity} {Resolution} with
  {Multi}-{Focal} {Attention}}. In \bibinfo{booktitle}{\emph{Proceedings of the
  54th {Annual} {Meeting} of the {Association} for {Computational}
  {Linguistics} ({Volume} 1: {Long} {Papers})}}.
  \bibinfo{publisher}{Association for Computational Linguistics},
  \bibinfo{address}{Berlin, Germany}, \bibinfo{pages}{621--631}.
\newblock
\urldef\tempurl%
\url{http://www.aclweb.org/anthology/P16-1059}
\showURL{%
\tempurl}


\bibitem[\protect\citeauthoryear{Hoffart, Yosef, Bordino, Furstenau, Pinkal,
  Spaniol, Taneva, Thater, and Weikum}{Hoffart et~al\mbox{.}}{2011}]%
        {hoffart_robust_nodate}
\bibfield{author}{\bibinfo{person}{Johannes Hoffart},
  \bibinfo{person}{Mohamed~Amir Yosef}, \bibinfo{person}{Ilaria Bordino},
  \bibinfo{person}{Hagen Furstenau}, \bibinfo{person}{Manfred Pinkal},
  \bibinfo{person}{Marc Spaniol}, \bibinfo{person}{Bilyana Taneva},
  \bibinfo{person}{Stefan Thater}, {and} \bibinfo{person}{Gerhard Weikum}.}
  \bibinfo{year}{2011}\natexlab{}.
\newblock \showarticletitle{Robust {Disambiguation} of {Named} {Entities} in
  {Text}}.
\newblock  (\bibinfo{year}{2011}), \bibinfo{pages}{11}.
\newblock


\bibitem[\protect\citeauthoryear{Kingma and Ba}{Kingma and Ba}{2014}]%
        {KingmaB14}
\bibfield{author}{\bibinfo{person}{Diederik~P. Kingma} {and}
  \bibinfo{person}{Jimmy Ba}.} \bibinfo{year}{2014}\natexlab{}.
\newblock \showarticletitle{Adam: {A} Method for Stochastic Optimization}.
\newblock \bibinfo{journal}{\emph{CoRR}}  \bibinfo{volume}{abs/1412.6980}
  (\bibinfo{year}{2014}).
\newblock
\urldef\tempurl%
\url{http://arxiv.org/abs/1412.6980}
\showURL{%
\tempurl}


\bibitem[\protect\citeauthoryear{Kipf, Fetaya, Wang, Welling, and Zemel}{Kipf
  et~al\mbox{.}}{2018}]%
        {kipf_neural_2018}
\bibfield{author}{\bibinfo{person}{Thomas Kipf}, \bibinfo{person}{Ethan
  Fetaya}, \bibinfo{person}{Kuan-Chieh Wang}, \bibinfo{person}{Max Welling},
  {and} \bibinfo{person}{Richard Zemel}.} \bibinfo{year}{2018}\natexlab{}.
\newblock \showarticletitle{Neural Relational Inference for Interacting
  Systems}.
\newblock  (\bibinfo{year}{2018}).
\newblock
\showeprint[arxiv]{1802.04687}
\urldef\tempurl%
\url{http://arxiv.org/abs/1802.04687}
\showURL{%
\tempurl}


\bibitem[\protect\citeauthoryear{Kipf and Welling}{Kipf and Welling}{2016}]%
        {KipfW16}
\bibfield{author}{\bibinfo{person}{Thomas~N. Kipf} {and} \bibinfo{person}{Max
  Welling}.} \bibinfo{year}{2016}\natexlab{}.
\newblock \showarticletitle{Semi-Supervised Classification with Graph
  Convolutional Networks}.
\newblock \bibinfo{journal}{\emph{CoRR}}  \bibinfo{volume}{abs/1609.02907}
  (\bibinfo{year}{2016}).
\newblock
\urldef\tempurl%
\url{http://arxiv.org/abs/1609.02907}
\showURL{%
\tempurl}


\bibitem[\protect\citeauthoryear{Klang and Nugues}{Klang and Nugues}{2014}]%
        {klang_named_nodate}
\bibfield{author}{\bibinfo{person}{Marcus Klang} {and} \bibinfo{person}{Pierre
  Nugues}.} \bibinfo{year}{2014}\natexlab{}.
\newblock \showarticletitle{Named {Entity} {Disambiguation} in a {Question}
  {Answering} {System}}.
\newblock  (\bibinfo{year}{2014}), \bibinfo{pages}{3}.
\newblock


\bibitem[\protect\citeauthoryear{Marcheggiani and Titov}{Marcheggiani and
  Titov}{2017}]%
        {marcheggiani_encoding_2017}
\bibfield{author}{\bibinfo{person}{Diego Marcheggiani} {and}
  \bibinfo{person}{Ivan Titov}.} \bibinfo{year}{2017}\natexlab{}.
\newblock \showarticletitle{Encoding {Sentences} with {Graph} {Convolutional}
  {Networks} for {Semantic} {Role} {Labeling}}. In
  \bibinfo{booktitle}{\emph{Proceedings of the 2017 {Conference} on {Empirical}
  {Methods} in {Natural} {Language} {Processing}}}.
  \bibinfo{publisher}{Association for Computational Linguistics},
  \bibinfo{address}{Copenhagen, Denmark}, \bibinfo{pages}{1506--1515}.
\newblock
\urldef\tempurl%
\url{https://www.aclweb.org/anthology/D17-1159}
\showURL{%
\tempurl}


\bibitem[\protect\citeauthoryear{McNamee and Dang}{McNamee and Dang}{2009}]%
        {mcnamee_overview_2009}
\bibfield{author}{\bibinfo{person}{Paul McNamee} {and}
  \bibinfo{person}{Hoa~Trang Dang}.} \bibinfo{year}{2009}\natexlab{}.
\newblock \showarticletitle{Overview of the {TAC} 2009 {Knowledge} {Base}
  {Population} {Track}}.
\newblock  (\bibinfo{year}{2009}).
\newblock


\bibitem[\protect\citeauthoryear{Meij, Balog, and Odijk}{Meij
  et~al\mbox{.}}{2014}]%
        {meij2014}
\bibfield{author}{\bibinfo{person}{Edgar Meij}, \bibinfo{person}{Krisztian
  Balog}, {and} \bibinfo{person}{Daan Odijk}.} \bibinfo{year}{2014}\natexlab{}.
\newblock \showarticletitle{Entity linking and retrieval for semantic search}.
  In \bibinfo{booktitle}{\emph{Proceedings of the {Seventh} {ACM}
  {International} {Conference} on {Web} {Search} and {Data} {Mining} - {WSDM}
  '14}}. \bibinfo{pages}{683--684}.
\newblock


\bibitem[\protect\citeauthoryear{Milne and Witten}{Milne and Witten}{2008}]%
        {milne_learning_2008}
\bibfield{author}{\bibinfo{person}{David Milne} {and} \bibinfo{person}{Ian~H.
  Witten}.} \bibinfo{year}{2008}\natexlab{}.
\newblock \showarticletitle{Learning to link with wikipedia}.
  \bibinfo{publisher}{ACM Press}, \bibinfo{pages}{509}.
\newblock
\showISBNx{978-1-59593-991-3}
\urldef\tempurl%
\url{https://doi.org/10.1145/1458082.1458150}
\showDOI{\tempurl}


\bibitem[\protect\citeauthoryear{Pappu, Blanco, Mehdad, Stent, and
  Thadani}{Pappu et~al\mbox{.}}{2017}]%
        {Pappu:2017:LME:3018661.3018724}
\bibfield{author}{\bibinfo{person}{Aasish Pappu}, \bibinfo{person}{Roi Blanco},
  \bibinfo{person}{Yashar Mehdad}, \bibinfo{person}{Amanda Stent}, {and}
  \bibinfo{person}{Kapil Thadani}.} \bibinfo{year}{2017}\natexlab{}.
\newblock \showarticletitle{Lightweight Multilingual Entity Extraction and
  Linking}. In \bibinfo{booktitle}{\emph{Proceedings of the Tenth ACM
  International Conference on Web Search and Data Mining}}
  \emph{(\bibinfo{series}{WSDM '17})}. \bibinfo{publisher}{ACM},
  \bibinfo{address}{New York, NY, USA}, \bibinfo{pages}{365--374}.
\newblock
\showISBNx{978-1-4503-4675-7}
\urldef\tempurl%
\url{https://doi.org/10.1145/3018661.3018724}
\showDOI{\tempurl}


\bibitem[\protect\citeauthoryear{Raiman and Raiman}{Raiman and Raiman}{2018}]%
        {raiman_deeptype:_2018}
\bibfield{author}{\bibinfo{person}{Jonathan Raiman} {and}
  \bibinfo{person}{Olivier Raiman}.} \bibinfo{year}{2018}\natexlab{}.
\newblock \showarticletitle{{DeepType}: {Multilingual} {Entity} {Linking} by
  {Neural} {Type} {System} {Evolution}}.
\newblock \bibinfo{journal}{\emph{arXiv:1802.01021 [cs]}} (\bibinfo{date}{Feb.}
  \bibinfo{year}{2018}).
\newblock
\urldef\tempurl%
\url{http://arxiv.org/abs/1802.01021}
\showURL{%
\tempurl}
\newblock
\shownote{arXiv: 1802.01021.}


\bibitem[\protect\citeauthoryear{Ren, Wu, He, Qu, Voss, Ji, Abdelzaher, and
  Han}{Ren et~al\mbox{.}}{2017}]%
        {ren_cotype:_2017}
\bibfield{author}{\bibinfo{person}{Xiang Ren}, \bibinfo{person}{Zeqiu Wu},
  \bibinfo{person}{Wenqi He}, \bibinfo{person}{Meng Qu},
  \bibinfo{person}{Clare~R. Voss}, \bibinfo{person}{Heng Ji},
  \bibinfo{person}{Tarek~F. Abdelzaher}, {and} \bibinfo{person}{Jiawei Han}.}
  \bibinfo{year}{2017}\natexlab{}.
\newblock \showarticletitle{{CoType}: {Joint} {Extraction} of {Typed}
  {Entities} and {Relations} with {Knowledge} {Bases}}. \bibinfo{publisher}{ACM
  Press}, \bibinfo{pages}{1015--1024}.
\newblock
\showISBNx{978-1-4503-4913-0}
\urldef\tempurl%
\url{https://doi.org/10.1145/3038912.3052708}
\showDOI{\tempurl}


\bibitem[\protect\citeauthoryear{Schlichtkrull, Kipf, Bloem, Berg, Titov, and
  Welling}{Schlichtkrull et~al\mbox{.}}{2017}]%
        {schlichtkrull_modeling_2017}
\bibfield{author}{\bibinfo{person}{Michael Schlichtkrull},
  \bibinfo{person}{Thomas~N. Kipf}, \bibinfo{person}{Peter Bloem},
  \bibinfo{person}{Rianne van~den Berg}, \bibinfo{person}{Ivan Titov}, {and}
  \bibinfo{person}{Max Welling}.} \bibinfo{year}{2017}\natexlab{}.
\newblock \showarticletitle{Modeling {Relational} {Data} with {Graph}
  {Convolutional} {Networks}}.
\newblock \bibinfo{journal}{\emph{arXiv:1703.06103 [cs, stat]}}
  (\bibinfo{date}{March} \bibinfo{year}{2017}).
\newblock
\urldef\tempurl%
\url{http://arxiv.org/abs/1703.06103}
\showURL{%
\tempurl}
\newblock
\shownote{arXiv: 1703.06103.}


\bibitem[\protect\citeauthoryear{Schuhmacher and Ponzetto}{Schuhmacher and
  Ponzetto}{2014}]%
        {schuhmacher_knowledge-based_2014}
\bibfield{author}{\bibinfo{person}{Michael Schuhmacher} {and}
  \bibinfo{person}{Simone~Paolo Ponzetto}.} \bibinfo{year}{2014}\natexlab{}.
\newblock \showarticletitle{Knowledge-based graph document modeling}.
  \bibinfo{publisher}{ACM Press}, \bibinfo{pages}{543--552}.
\newblock
\showISBNx{978-1-4503-2351-2}
\urldef\tempurl%
\url{https://doi.org/10.1145/2556195.2556250}
\showDOI{\tempurl}


\bibitem[\protect\citeauthoryear{Usbeck, Ngonga~Ngomo, Röder, Gerber, Coelho,
  Auer, and Both}{Usbeck et~al\mbox{.}}{2014}]%
        {mika_agdistis_2014}
\bibfield{author}{\bibinfo{person}{Ricardo Usbeck},
  \bibinfo{person}{Axel-Cyrille Ngonga~Ngomo}, \bibinfo{person}{Michael
  Röder}, \bibinfo{person}{Daniel Gerber}, \bibinfo{person}{Sandro~Athaide
  Coelho}, \bibinfo{person}{Sören Auer}, {and} \bibinfo{person}{Andreas
  Both}.} \bibinfo{year}{2014}\natexlab{}.
\newblock \showarticletitle{{AGDISTIS} - {Graph}-{Based} {Disambiguation} of
  {Named} {Entities} {Using} {Linked} {Data}}.
\newblock In \bibinfo{booktitle}{\emph{The {Semantic} {Web} – {ISWC} 2014}},
  \bibfield{editor}{\bibinfo{person}{Peter Mika}, \bibinfo{person}{Tania
  Tudorache}, \bibinfo{person}{Abraham Bernstein}, \bibinfo{person}{Chris
  Welty}, \bibinfo{person}{Craig Knoblock}, \bibinfo{person}{Denny
  Vrandečić}, \bibinfo{person}{Paul Groth}, \bibinfo{person}{Natasha Noy},
  \bibinfo{person}{Krzysztof Janowicz}, {and} \bibinfo{person}{Carole Goble}}
  (Eds.). Vol.~\bibinfo{volume}{8796}. \bibinfo{publisher}{Springer
  International Publishing}, \bibinfo{address}{Cham},
  \bibinfo{pages}{457--471}.
\newblock
\showISBNx{978-3-319-11963-2 978-3-319-11964-9}
\urldef\tempurl%
\url{https://doi.org/10.1007/978-3-319-11964-9_29}
\showDOI{\tempurl}


\bibitem[\protect\citeauthoryear{Vrande\v{c}i\'{c}}{Vrande\v{c}i\'{c}}{2012}]%
        {Vrandecic2012}
\bibfield{author}{\bibinfo{person}{Denny Vrande\v{c}i\'{c}}.}
  \bibinfo{year}{2012}\natexlab{}.
\newblock \showarticletitle{Wikidata: A New Platform for Collaborative Data
  Collection}. In \bibinfo{booktitle}{\emph{Proceedings of the 21st
  International Conference on World Wide Web}} \emph{(\bibinfo{series}{WWW '12
  Companion})}. \bibinfo{publisher}{ACM}, \bibinfo{address}{New York, NY, USA},
  \bibinfo{pages}{1063--1064}.
\newblock
\showISBNx{978-1-4503-1230-1}
\urldef\tempurl%
\url{https://doi.org/10.1145/2187980.2188242}
\showDOI{\tempurl}


\end{thebibliography}


\appendix
\section{Configurations} 
\label{app:configuration}
~\\~\\
\begin{table}[h!]
    \centering
    \caption*{Feedforward of averages}
    \begin{tabular}{ c c }
        \toprule
            \multicolumn{1}{c}{\textbf{Parameter}} & \multicolumn{1}{c}{\textbf{Value}} \\
        \midrule
            Final hidden layer size  & $250$ dim \\ 
            Output size  & $2$ dim \\ 
        \midrule
            $\mathbf{y}_\mathrm{text}$ size   & $300$ dim \\
        	$\mathbf{y}_\mathrm{graph}$ size   & $300$ dim \\
            Batch size               & 10\\
        \bottomrule
    \end{tabular}
\end{table}

\begin{table}[h!]
    \centering
    \caption*{Text \ac{LSTM}+centroid}
    \begin{tabular}{ c c }
        \toprule
            \multicolumn{1}{c}{\textbf{Parameter}} & \multicolumn{1}{c}{\textbf{Value}} \\
        \midrule
            Word vectors size & $300$ dim \\
            Final hidden layer size  & $250$ dim \\ 
            Output size  & $2$ dim \\ 
            Dense layer before text \ac{LSTM} & $50$ dim \\
            Text \ac{LSTM} memory         & ($2\times$) $100$ dim \\
            $\mathbf{y}_\mathrm{text}$ size    & $150$ dim \\
            $\mathbf{y}_\mathrm{graph}$ size   & $300$ dim \\
            Batch size               & 10\\
        \bottomrule
    \end{tabular}
\end{table}

\begin{table}[h!]
    \centering
    \caption*{Text \ac{LSTM}+\ac{RNN} of nodes}
    \begin{tabular}{ c c }
        \toprule
            \multicolumn{1}{c}{\textbf{Parameter}} & \multicolumn{1}{c}{\textbf{Value}} \\
            Word vectors size & $300$ dim \\
            Node vectors size & $300$ dim \\
            Final hidden layer size  & $250$ dim \\ 
            Output size  & $2$ dim \\ 
            dense layer before text \ac{LSTM} & $50$ dim \\
            Dext \ac{LSTM} memory         & ($2\times$) $100$ dim \\
            $\mathbf{y}_\mathrm{text}$ size    & $150$ dim \\
            $\mathbf{y}_\mathrm{graph}$ size   & $250$ dim \\
            Batch size               & 10\\
        \bottomrule
    \end{tabular}
\end{table}

\begin{table}[h!]
    \centering
    \caption*{Text \ac{LSTM}+\ac{RNN} of triplets}
    \begin{tabular}{ c c }
        \toprule
            \multicolumn{1}{c}{\textbf{Parameter}} & \multicolumn{1}{c}{\textbf{Value}} \\
            Word vectors size & $300$ dim \\
            Triplet vectors size & $1200$ dim \\
            Final hidden layer size  & $250$ dim \\ 
            Output size  & $2$ dim \\ 
            Dense layer before text \ac{LSTM} & $50$ dim \\
            Text \ac{LSTM} memory         & ($2\times$) $100$ dim \\
            Dense layer before graph \ac{LSTM} & $50$ dim \\
            Graph \ac{LSTM} memory         & ($2\times$) $100$ dim \\
            $\mathbf{y}_\mathrm{text}$ size    & $150$ dim \\
            $\mathbf{y}_\mathrm{graph}$ size   & $250$ dim \\
            Batch size               & 10\\
        \bottomrule
    \end{tabular}
\end{table}

\begin{table}[h!]
    \centering
    \caption*{Text \ac{LSTM}+\ac{GCN}}
    \begin{tabular}{ c c }
        \toprule
            \multicolumn{1}{c}{\textbf{Parameter}} & \multicolumn{1}{c}{\textbf{Value}} \\
            Word vectors size & $300$ dim \\
            Node (and edges) vectors size & $300$ dim \\
            Final hidden layer size  & $250$ dim \\ 
            Output size  & $2$ dim \\ 
            Dense layer before text \ac{LSTM} & $50$ dim \\
            Text \ac{LSTM} memory         & ($2\times$) $100$ dim \\
            $\mathbf{y}_\mathrm{text}$ size    & $150$ dim \\
            $\mathbf{y}_\mathrm{graph}$ size   & $250$ dim \\
            \ac{GCN} layer           & $250$ dim \\
            Latent attention dimension $d$ & $250$ dim \\
            Edge Dropout             & $0.9$ (\emph{keep probability}) \\
            Batch size               & 10\\
        \bottomrule
    \end{tabular}
\end{table}


\end{document}